\documentclass[11pt]{article}
\usepackage{coling2016}
\usepackage{times}
\usepackage{url}
\usepackage{latexsym}
\usepackage{amssymb}
\usepackage{amsmath}
\usepackage{booktabs}
\usepackage[utf8]{inputenc}
\usepackage{covington}

\usepackage{todonotes}

\title{Byte-based Language Identification with Deep Convolutional Networks}

\author{Johannes Bjerva \\
  University of Groningen \\
  The Netherlands \\
  {\tt j.bjerva@rug.nl} \\
}

\date{}

\begin{document}
\maketitle
\begin{abstract}
We report on our system for the shared task on discrimination of similar languages (DSL 2016).
The system uses only byte representations in a deep residual network (ResNet).
The system, named ResIdent, is trained only on the data released with the task (closed training).
We obtain 84.88\% accuracy on subtask A, 68.80\% accuracy on subtask B1, and 69.80\% accuracy on subtask B2.
A large difference in accuracy on development data can be observed with relatively minor changes in our network's architecture and hyperparameters.
We therefore expect fine-tuning of these parameters to yield higher accuracies.
\end{abstract}

\section{Introduction}

Language identification is an unsolved problem, certainly in the context of discriminating between very similar languages \cite{baldwin:2010}.
This problem is tackled in the Discriminating between Similar Languages (DSL) series of shared tasks \cite{zampieri:2014:VarDial,zampieri:2015:LT4VarDial}.
Most successful approaches to the DSL shared task in previous years have relied on settings containing ensembles of classifiers \cite{dslrec:2016}.
These classifiers often use various combinations of features, mostly based on word, character, and/or byte $n$-grams (see, e.g., Cavnar et al.~\shortcite{cavnar:1994}, Lui and Baldwin~\shortcite{lui:2012}).

We are interested in exploring a single methodological aspect in the current edition of this shared task \cite{dsl2016}.
We aim to investigate whether reasonable results for this task could be obtained by applying recently emerged neural network architectures, coupled with sub-token input representations.
To address this question, we explore convolutional neural
networks (CNNs) and recurrent neural networks (RNNs).
Deep residual networks (ResNets) are a recent building block for CNNs which have yielded promising results in, e.g.,  image classification tasks \cite{resnets:2015,resnets:2016}.
ResNets are constructed by stacking so-called residual units.
These units can be viewed as a series of convolutional layers with a `shortcut' which facilitates signal propagation in the neural network.
This, in turn, allows for training deeper networks more easily \cite{resnets:2016}.
In Natural Language Processing (NLP), ResNets have shown state-of-the-art performance for Semantic and Part-of-Speech tagging \cite{bjerva:2016:coling}.
However, no previous work has attempted to apply ResNets to language identification.


\section{Method}

\begin{figure}[h]
    \centering
    \includegraphics[width=0.5\textwidth]{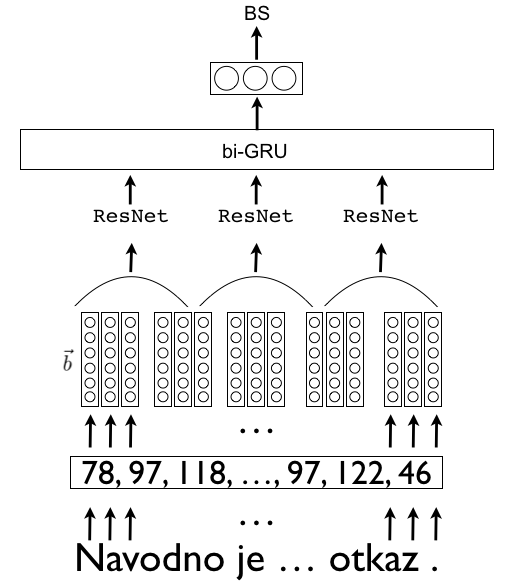}
    \caption{\label{fig:model_arch}Model architecture: ResNet with byte representations ($\vec{b}$), with a bi-GRU at the upper level. The input example sequence is converted to a sequence of byte identifiers (one integer per byte, rather than one integer per character), which are converted to a byte embedding representation. This input is treated by the ResNet, followed by the bi-GRU, finally yielding the language id \textit{BS} (Bosnian).}
\end{figure}

Several previous approaches in the DSL shared tasks have formulated the task as a two-step classification, first identifying the language group, and then the specific language \cite{zampieri:2015:LT4VarDial}.
Instead of taking this approach, we formulate the task as a multi-class classification problem, with each language / dialect representing a separate class.
Our system is a deep neural network consisting of a bidirectional Gated Recurrent Unit (GRU) network at the upper level, and a Deep Residual Network (ResNet) at the lower level (Figure~\ref{fig:model_arch}).
The inputs of our system are byte-level representations of each input sentence, with byte embeddings which are learnt during training.
Using byte-level representations differs from character-level representations in that UTF-8 encodes non-ascii symbols with more than one byte, which potentially allows for more disambiguating power.
A concrete example can be found when considering the relatively similar languages Norwegian and Swedish.
Here, there are two pairs of letters which are interchangeable: where Swedish uses `ä' (C3 A4) and `ö' (C3 B6), Norwegian uses `æ' (C3 A6) and `ø' (C3 B8).
Hence, using the lower-level byte representation, we allow the model to take advantage of the first shared byte between these characters.
The architecture used in this work is based on the sequence-to-sequence labelling architecture used in Bjerva et al.~\shortcite{bjerva:2016:coling}, modified for the task of language identification.
Our system is implemented in Keras using the Tensorflow backend \cite{keras,tensorflow}.

\subsection{Gated Recurrent Unit Networks}
GRUs~\cite{cho:ea:2014} are a recently introduced variant of RNNs, and are designed to prevent vanishing gradients, thus being able to cope with longer input sequences than vanilla RNNs.
GRUs are similar to the more commonly-used Long Short-Term Memory networks (LSTMs), both in purpose and implementation \cite{gru}.
A bi-directional GRU makes both forward and backward passes over sequences, and can therefore use both preceding and succeeding contexts to predict a tag \cite{graves:schmidhuber:2005,goldberg:primer}.
Bi-directional GRUs and LSTMs have been shown to yield high performance on several NLP tasks, such as POS and semantic tagging, named entity tagging, and chunking \cite{wang2015:unified,yang:2016,plank:2016,bjerva:2016:coling}.

\subsection{Deep Residual Networks}
\label{sec:resnets}
Deep Residual Networks (ResNets) are built up by stacking residual units.
A residual unit can be expressed as:
\begin{equation}
    \begin{aligned}
    &y_l = h(x_l) + \mathcal{F}(x_l,\mathcal{W}_l),\\
    &x_{l+1} = f(y_l),
    \end{aligned}
\end{equation}
\noindent where $x_l$ and $x_{l+1}$ are the input and output of the $l$-th layer, $\mathcal{W}_l$ is the weights for the $l$-th layer, and $\mathcal{F}$ is a residual function \cite{resnets:2016}, e.g., the identity function \cite{resnets:2015}, which we also use in our experiments.
ResNets can be intuitively understood by thinking of residual functions as paths through which information can propagate easily.
This means that, in every layer, a ResNet learns more complex feature combinations, which it combines with the shallower representation from the previous layer.
This architecture allows for the construction of much deeper networks.
ResNets have recently been found to yield impressive performance in both image recognition and NLP tasks \cite{resnets:2015,resnets:2016,robert:sigmorphon:2016,conneau:2016}, and are an interesting and effective alternative to simply stacking layers.
In this paper we use the \textit{assymetric} variant of ResNets as described in Equation 9 in He et al.~\shortcite{resnets:2016}:
\begin{equation}
    \begin{aligned}
    &x_{l+1} = x_l + \mathcal{F}(\hat{f}(x_l), \mathcal{W}_l).
    \end{aligned}
\end{equation}

Our residual block, using dropout and batch normalization \cite{dropout,batchnorm}, is defined in Table~\ref{tab:resblock}.
In the table, \textit{merge} indicates the concatenation of the input of the residual block, with the output of the final convolutional layer.
\vspace{-0.15cm}
\begin{table}[htbp]
    \centering
    \begin{tabular}{lrr}
    \toprule
    \textbf{type} & \textbf{patch/pool size} \\
    \midrule
    Batch normalization + ReLu + Dropout ($p=0.5$) & \\
    Convolution & $8$ \\
    Batch normalization + ReLu + Dropout ($p=0.5$) & \\
    Convolution & $4$ \\
    Merge  & \\
    Maximum pooling & $2$ \\
    \bottomrule
    \end{tabular}
\caption{\label{tab:resblock}Residual block overview.}
\end{table}
\vspace{-0.3cm}

\subsection{System Description}

We represent each sentence using a byte-based representation ($S_b$).
This representation is a 2-dimensional matrix $S_b \in \mathbb{R}^{s \times d_b}$, where $s$ is the zero-padded sentence length and $d_b$ is the dimensionality of the byte embeddings.
Byte embeddings are first passed through a ResNet in order to obtain a representation which captures something akin to byte $n$-gram features.\footnote{Note that bytes are passed through the ResNet \textit{one by one}, yielding one representation per byte, rather than as a whole sequence, which would yield a single representation per sentence.}
The size of $n$ is determined by the convolutional window size used.
We use a convolutional window size with length $8$, meaning that for each byte in the input, the ResNet can learn a suitable representation incorporating up to $8$ bytes of context information.
These overlapping byte-based $n$-gram features are then passed through to the bi-GRU, which yields a \textit{sentence level} representation.
The softmax layer applied to the bi-GRU output is then used in order to obtain the network's predicted class per input.

\subsubsection{Hyperparameters}

The hyperparameters used by the system were tuned on an altogether different task (semantic tagging), and adapted for the current task.
The dimensionality of our byte embeddings, $d_b$, is set to $64$.
Our residual block is defined in Section~\ref{sec:resnets}.
We use rectified linear units (ReLUs) for all activation functions \cite{relu}, and apply dropout with $p=0.1$ to both input weights and recurrent weights in the bi-GRU.
All GRU layers have $100$ hidden units.

All experiments were run with early stopping monitoring validation set loss, using a maximum of 50 epochs, and a batch size of 100.
Optimisation is done using the ADAM algorithm \cite{adam}, with the categorical cross-entropy loss function as training objective.

For the B tasks, we train the model in the same way as for the A tasks.
Only a handful of instances ($n\approx5$) per B run are classified as belonging to a language which the B group does not contain.
These cases are automatically converted to being in the class \textit{hr}.
For the B tasks, we also perform a simple clean-up of the data.
We first remove all hyperlinks, hashtags and usernames from the text with a simple regex-based script.
We then remove all tweets classified as English.
We submitted three runs for each subtask.
The system used for runs 1, 2 and 3 contain five, four and three residual blocks respectively.

\section{Results}
\label{sec:results}

\begin{table}[htbp]
\center
\begin{tabular}{lllllll}
\toprule
\bf Test Set & \bf Track & \bf Run & \bf Accuracy & \bf F1 (micro) & \bf F1 (macro) & \bf F1 (weighted) \\
\midrule
A & closed & Baseline & 0.083 & & & \\
A & closed & run1 & 0.8462 & 0.8462 & 0.8415 & 0.8415 \\
A & closed & run2 & 0.8324 & 0.8324 & 0.8272 & 0.8272 \\
A & closed & run3 & \bf 0.8488 & 0.8488 & 0.8467 & 0.8467 \\
\midrule
B1 & closed & Baseline & 0.020 & & & \\
B1 & closed & run1 & 0.682 & 0.682 & 0.6802 & 0.6802 \\
B1 & closed & run2 & 0.676 & 0.676 & 0.6708 & 0.6708 \\
B1 & closed & run3 & \bf 0.688 & 0.688 & 0.6868 & 0.6868 \\
\midrule
B2 & closed & Baseline & 0.020 & & & \\
B2 & closed & run1 & 0.684 & 0.684 & 0.6788 & 0.6788 \\
B2 & closed & run2 & \bf 0.698 & 0.698 & 0.6942 & 0.6942 \\
B2 & closed & run3 & 0.664 & 0.664 & 0.6524 & 0.6524 \\
\bottomrule
\end{tabular}
\caption{\label{tab:results-all}Results for all runs in subtasks A, B1 and B2 (closed training).}
\end{table}

\begin{table}[htbp]
    \centering
    \begin{tabular}{lrrrrrrrrrrrr}
    \toprule
           &   es-ar &   es-es &   es-mx &   fr-ca &   fr-fr &    id &   my &   pt-br &   pt-pt &  hr &  bs &    sr \\
    \midrule
     es-ar &     824 &      77 &      94 &       0 &       1 &     1 &    0 &       2 &       1 &   0 &   0 &     0 \\
     es-es &      90 &     778 &     127 &       0 &       1 &     0 &    0 &       1 &       2 &   0 &   1 &     0 \\
     es-mx &     210 &     269 &     520 &       0 &       0 &     0 &    0 &       1 &       0 &   0 &   0 &     0 \\
     \midrule
     fr-ca &       0 &       0 &       0 &     956 &      44 &     0 &    0 &       0 &       0 &   0 &   0 &     0 \\
     fr-fr &       0 &       0 &       0 &      93 &     905 &     0 &    0 &       1 &       0 &   1 &   0 &     0 \\
     \midrule
     id    &       0 &       0 &       0 &       0 &       0 &   951 &   48 &       0 &       0 &   0 &   0 &     1 \\
     my    &       0 &       0 &       0 &       0 &       0 &    30 &  970 &       0 &       0 &   0 &   0 &     0 \\
     \midrule
     pt-br &       0 &       0 &       1 &       0 &       0 &     0 &    0 &     891 &     107 &   1 &   0 &     0 \\
     pt-pt &       0 &       1 &       0 &       0 &       0 &     0 &    0 &      78 &     920 &   0 &   1 &     0 \\
     \midrule
     hr    &       0 &       0 &       0 &       0 &       0 &     0 &    0 &       0 &       0 & 823 & 150 &    27 \\
     bs    &       0 &       0 &       0 &       0 &       1 &     0 &    0 &       1 &       0 & 143 & 730 &   125 \\
     sr    &       0 &       0 &       0 &       0 &       1 &     0 &    0 &       0 &       0 &  15 &  67 &   917 \\
    \bottomrule
    \end{tabular}
    \caption{\label{tab:conf_mat_A}Confusion matrix, closed run 3, on test set A. The x-axis indicates predicted labels, and the y-axis indicates true labels.}
\end{table}

\begin{table}[htbp]
    \centering
    \begin{tabular}{lrrrrr|rrrrr}
    \toprule
    & & \textbf{B1} & & & & & & \textbf{B2} & & \\
    \midrule
           &   pt-br &   pt-pt & hr &    bs &    sr  &  pt-br &   pt-pt & hr &   bs &    sr \\
    \midrule
     pt-br &     74 &      24 &  1 &     0 &     1 &     54 &      40 &  3 &    2 &     1 \\
     pt-pt &     31 &      67 &  1 &     0 &     1 &     15 &      80 &  5 &    0 &     0 \\
     \midrule
     bs    &      0 &       0 & 60 &    31 &     9 &      0 &       0 & 75 &   20 &     5 \\
     hr    &      1 &       0 & 20 &    62 &    17 &      0 &       0 & 31 &   56 &    13 \\
     sr    &      4 &       0 &  5 &    10 &    81 &      2 &       0 &  8 &    6 &    84 \\

    \bottomrule
    \end{tabular}
\caption{\label{tab:conf_mat_B}Confusion matrix, closed run 3 on test set B1 (left) and closed run 2 on test set B2 (right). The x-axis indicates predicted labels, and the y-axis indicates true labels.}
\end{table}

We evaluate our system in subtasks A, B1 and B2.
Subtask A contains data for five language groups, with two to three languages in each group \cite{tan:2014:BUCC}.
Subtask B1 and B2 contain data for a subset of the languages in subtask A, compiled from Twitter.
Subtask B1 contains the amount of tweets necessary for a human annotator to make reliable judgements, whereas B2 contains the maximum amount of data available per tweet.

For subtasks A and B1, run 3 results in the best accuracy on test, whereas run 2 results in the best accuracy on B2.
The results are shown in Table~\ref{tab:results-all}.
Table~\ref{tab:conf_mat_A} and Table~\ref{tab:conf_mat_B} contain confusion matrices for the results in subtask A and B respectively.

\section{Discussion}

Judging from the confusion matrices in Section~\ref{sec:results}, our system has very low confusion between language groups.
However, confusion can be observed within all groups.
Although the system achieves reasonable performance, there is a large gap between our system and the best performing systems (e.g. \c{C}\"{o}ltekin and Rama~\shortcite{coltekin:2016}, who obtain 89.38\% accuracy on task A, 86.2\% on B1, and 82.2\% on B2).
This can to some extent be explained by limitations caused by our implementation.

The largest limiting factor can be found in the fact that we only allowed our system to use the first ca. 384 bytes of each training/testing instance.
For the training and development set, and subtask A, this was no major limitation, as this allowed us to use more than 90\% of the available data.
However, for subtasks B1 and B2, this may have seriously affected the system's performance.
Additionally, we restricted our system to using only byte embeddings as input.
Adding word-level representations into the mix, would likely increase system performance.

We also observed considerable differences in development accuracy when changing hyperparameters of our network in relatively minor ways.
For instance, altering the patch sizes used in our CNNs had a noticeable impact on validation loss.
However, altering the amount of residual blocks used, did not have a large effect on results.
The neural network architecture, as well as most of the hyperparameters, were tuned on an altogether different task (semantic tagging), and adapted for the current task.
Further fine tuning of the network architecture and hyperparameters for this task would therefore likely lead to narrowing the performance gap.

\section{Conclusions}

We implemented a language identification system using deep residual networks (ResNets) coupled with a bidirectional Gated Recurrent Unit network (bi-GRU), using only byte-level representations.
In the DSL 2016 shared task, we achieved reasonable performance, with 84.88\% accuracy on subtask A, 68.80\% accuracy on subtask B1, and 69.80\% accuracy on subtask B2.
Although acceptable performance was achieved, further fine tuning of input representations and system architecture would likely improve performance.

\section*{Acknowledgements}
We would like to thank the Center for Information Technology of the University of Groningen for their support and for providing access to the Peregrine high performance computing cluster, as well as the anonymous reviewers for their valuable feedback.
%

\bibliographystyle{acl}
\bibliography{comsem}

\end{document}